\documentclass{article} % For LaTeX2e
\usepackage{collas2022_conference,times}

\collasfinaltrue

\usepackage{hyperref}
\hypersetup{
    colorlinks=true,
    linkcolor=red,
    filecolor=magenta,      
    urlcolor=blue,
    citecolor=purple,
    pdftitle={Overleaf Example},
    pdfpagemode=FullScreen,
    }

\usepackage{amsmath,amsfonts,bm}

\def\eqref#1{equation~\ref{#1}}
\def\1{\bm{1}}

\DeclareMathAlphabet{\mathsfit}{\encodingdefault}{\sfdefault}{m}{sl}
\SetMathAlphabet{\mathsfit}{bold}{\encodingdefault}{\sfdefault}{bx}{n}

\DeclareMathOperator*{\argmin}{arg\,min}

\usepackage{graphicx} %[dvips]
\usepackage{comment}

\usepackage{amsmath}
\usepackage{amsthm}
\usepackage{amssymb}
\usepackage{amsfonts,bm,bbm}

\usepackage{wrapfig}
\usepackage{cleveref}
\usepackage[justification=justified]{subfig}
\usepackage{placeins}
\usepackage{booktabs}
\usepackage{blindtext}

\usepackage[ruled,linesnumbered]{algorithm2e}

\usepackage{tabularx}

\makeatletter
\def\th@plain{%
  \thm@notefont{}% same as heading font
  \itshape % body font
}
\def\th@definition{%
  \thm@notefont{}% same as heading font
  \normalfont % body font
}
\makeatother

\newtheorem{theorem}{Theorem}[section]

\newcommand{\hR}{\widehat{R}}

\newcommand{\reals}{\mathbb{R}}

\newcommand{\emprisk}{\hR}
\newcommand{\risk}{R}

\newcommand{\given}{\quad | \quad}

\newcommand{\spa}{\hspace{0.25em}}

\DeclareMathOperator*{\prob}{\mathbb{P}}
\DeclareMathOperator*{\ex}{\mathbb{E}}
\DeclareMathOperator*{\eqdef}{\triangleq}

\title{Test Sample Accuracy Scales with \\ Training Sample Density in Neural Networks}

\author{
	\hspace*{-0.5em} Xu Ji \\
	Mila \\
	\And
	Razvan Pascanu\\
	DeepMind \\
	\And 
	Devon Hjelm\\
	Mila, MSR \\
	\And
	Balaji Lakshminarayanan\\
	Google Brain \\
	\And
	Andrea Vedaldi \\
	Oxford University
}

\begin{document}

\maketitle

\begin{abstract}
Intuitively, one would expect accuracy of a trained neural network's prediction on test samples to correlate with how densely the samples are surrounded by seen training samples in representation space. We find that a bound on empirical training error smoothed across linear activation regions scales inversely with training sample density in representation space. Empirically, we verify this bound is a strong predictor of the inaccuracy of the network's prediction on test samples. For unseen test sets, including those with out-of-distribution samples, ranking test samples by their local region's error bound and discarding samples with the highest bounds raises prediction accuracy by up to 20\% in absolute terms for image classification datasets, on average over thresholds.
\end{abstract}

\section{Introduction}

When do trained models make mistakes? Intuitively, one expects higher prediction error for test samples that are more \emph{novel}, compared to seen training data.
For neural network inference models, one  measure of sample novelty is distance from training samples in representation space, according to some distance metric $k$.
Integrated over all training samples, this corresponds to the sample falling in a low density region in the metric space defined by $k$.
A number of existing works on detecting out-of-distribution samples relate to this idea. For instance \Citet{lee2018simple} and \citet{tack2020csi} use distance to estimated modes in network representation space as a measure of prediction unreliability.
In non-parametric Gaussian Process inference \citep{rasmussen2003gaussian}, prediction certainty corresponds exactly to local density of training samples.
The idea of this work is to derive and empirically test a similar measure of prediction unreliability for ReLU neural networks.

%Traditionally, the reliability of a function is characterized with generalization error bounds that provide probabilistic upper bounds on expected test error for any draw of empirical training data. 
Since the input space of a ReLU network can be partitioned into linear activation regions \citep{montufar2014number} such that each region is mapped to a distinct linear function that uses a different parameterization or subset of network weights - which for convenience we call the subfunctions of the network - 
one could cast the novelty or unreliability of a sample as a bound on error of the specific subfunction it induces.
Unlike in error bounding for model selection, the objective is to take a pre-trained model and rank test samples, so the model is fixed.
For deep networks, test samples typically fall in linear activation regions unpopulated by empirical training samples, so bounds that are a function of empirical training error are undefined.
We propose to smooth the empirical error by taking a weighted average of empirical errors across activation regions where the weighting is defined by a function $k$ of representation distance.
Constructing a bound on smooth empirical error yields a quantity that scales exactly with the inverse of density of training samples in the representation space defined by $k$.

%Why is the smooth empirical error function meaningful? 

There are many possible ways to partition a neural network. 
An argument against making the partitioning more coarse than linear activation regions is reduction in discriminativity; in the extreme case, casting the network as a single subfunction results in the same error bound for all samples.
An argument against making the partitioning more finegrained is linear functions are already the least complex class of parametric functions; there is no way to further subdivide a linear activation region such that the functions computed are different, as each input in the region implicates exactly the same parameters.
However, an interesting direction for future work is extending the idea of density-as-reliability to 
density in continuous space, rather than of discrete partitions, which would require a different bound to the one used in this work. 

\begin{figure}[h]
    \centering
    \subfloat[][]{\includegraphics[width=0.19\textwidth]{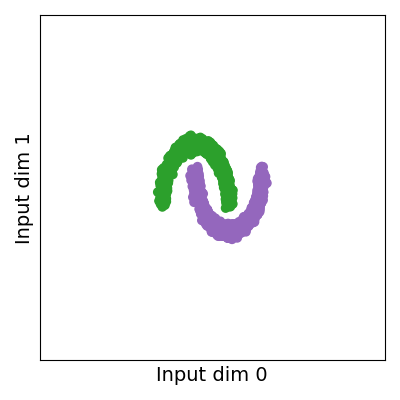}}\hfill
    \subfloat[][
    ]{\includegraphics[width=0.19\textwidth]{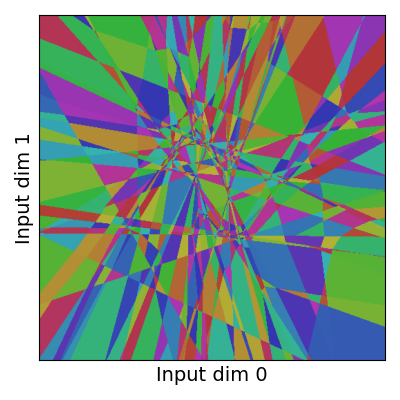}}\hfill
    \subfloat[][  ]{\includegraphics[width=0.22\textwidth]{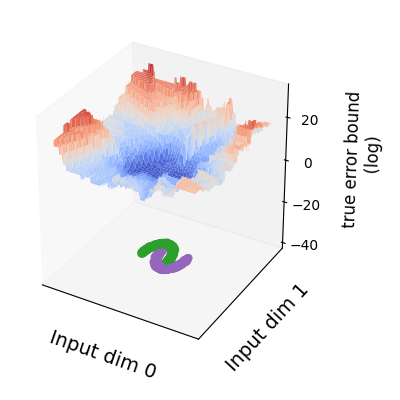}\label{f:two_moons_d}}\hfill
    \subfloat[][
    %Expressivity is highest around training data.
    ]{\includegraphics[width=0.24\textwidth]{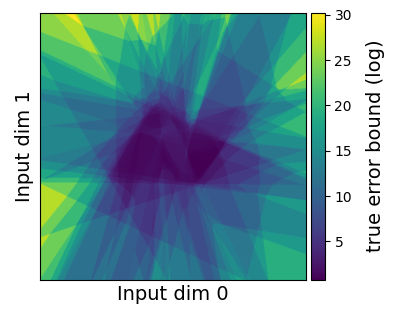}}\hfill
    \caption{
    Shattering a neural network trained on the halfmoons dataset into co-dependent linear subfunctions to obtain a heatmap of unreliability across the input space. The model is a MLP with two hidden ReLU layers of 32 units each. (a) Training data. (b) Linear activation regions. (c) Subfunction smooth error bound as unreliability. %The activation pattern considered all nodes from both layers. 
    (d) Unreliability heatmap in 2D.
    }
    \label{f:two_moons}
\end{figure}

\Cref{f:two_moons} illustrates our approach for a model trained on the half moons dataset.
%The network is shattered into subfunctions corresponding to linear activation regions, and a generalization error bound is derived that uses smooth training sample density. 
%Whereas model-level bounds produce one value for the model, 
Considering the model as a composition of individual functions allows the discrimination of risk based on input region. 

\section{Input-dependent unreliability}

\subsection{Notation}

% start with what they know, add subfunctions last!
Let $f : \mathcal{X} \rightarrow \mathcal{Y}$ be a piecewise linear neural network comprised of linear functions and ReLU activation functions. Let $V = (v_j)_{1 \leq j \leq M}$, ordered according to any fixed ordering, be the set of $M$ ReLUs in $f$ where $\forall j: v_j: \mathcal{X} \rightarrow \{0, 1\}$ is 1 if the ReLU is positive valued when $f(x)$ is computed, and 0 otherwise.
Let patterns set $P \subset \{0, 1\}^{M}$, $P = (p_i)_{1 \leq i \leq C}$ be the set of all feasible ReLU activation decisions, meaning all possible binary patterns induced by traversing the input space $\mathcal{X}$. In general this is not equal to the full bit space $\{0, 1\}^{M}$, and tractably computing it exactly is an open problem \citep{montufar2014number}. There is a bijection between these patterns and the linear activation regions of the network, as each linear activation region is uniquely determined by the set of ReLU activation decisions \citep{raghu2017expressive}.
%Recall that activation regions are the non-overlapping polytopes in input space corresponding to activation patterns. 
Let activation region $a_i$ correspond to pattern $p_i$:
\begin{align}
    a_i  \eqdef \{x \in \mathcal{X} \given \forall j \in [1, M]: p_{i}^j = v_j(x)\}, \label{eq:regions}
\end{align}
%Note that subfunctions are identified by the pattern
%and have a different domain,
%so subfunctions with different patterns are considered distinct even if behaviorally identical.
Each pattern $p_i$ defines a distinct linear function $h_i : \mathcal{X} \rightarrow \mathcal{Y}$, or \emph{subfunction}, by the replacement of the ReLU corresponding to each $v_j$ (non-linear) by the identity function if $p_i^j = 1$ or the zero function otherwise (linear). Let the set of all subfunctions be $H = (h_i)_{1 \leq i \leq C}$, ordered identically to $P$. 
Since activation regions are non-overlapping polytopes in input space, each input sample belongs to a unique activation region, so inference with model $f$ can be rewritten as $f(x) = h_i(x)$ where $x \in a_i$.
%That is, inference on a given input involves the simultaneous selection and computation of one subfunction. 
For convenience we overload the definition of $H$ so the subfunction induced for any input sample $x$ is $H(x) \eqdef h_i \text{ where } x \in a_i$. 

\subsection{Smooth empirical error}

Consider a training dataset $S$ containing $N$ sampled pairs, $S \in (\mathcal{X} \times \mathcal{Y})^N$, where samples are drawn iid from distribution $D$.
Define the empirical error or risk of the full network on this dataset:
\begin{align}
 \emprisk_S(f) \eqdef \frac{1}{|S|} \sum_{n=1}^{|S|} r(S^n; f), \label{eq:emp_risk}
\end{align}
where $r$ is a bounded error function on samples: $0 \le r((x, y); f) \le 1$ for all $x, y \in (\mathcal{X} \times \mathcal{Y})$ and $S^n$ is the $n$-th element of $S$.
Note empirical error, e.g. proportion of incorrect classifications, is not necessarily the objective function for training; it is fine for the latter to not have finite bounds.
Quantify the empirical error for a subfunction $h_i$ using standard empirical error (\cref{eq:emp_risk}). The activation region dataset is $S_i \eqdef S \cap a_i$ with $N_i \eqdef |S_i|$ samples drawn iid from its data distribution $D_i$, defined as $\prob_{D_i}(x, y) = \prob_D(x, y | x \in a_i)$. Then: 
%x,y \sim D_i \rightarrow 
%
\begin{align}
 \emprisk_{S_i}(h_i) &= \frac{1}{N_i} \sum_{n=1}^{N_i} r(S_i^n; f), \\
 \text{and for any $\delta \in (0, 1]$, with probability }&\text{$> 1 - \delta$:} \qquad \ex_{S_i} [\emprisk_{S_i}(h_i)] \leq \emprisk_{S_i}(h_i) + \sqrt{\frac{\log \frac{2}{\delta}}{2N_i}}. \label{eq:hyperlocal_bound}
\end{align}

\begin{wrapfigure}[18]{r}{0.40\textwidth}
%\vspace{-1em}
 \centering
 \includegraphics[width=0.40\textwidth]{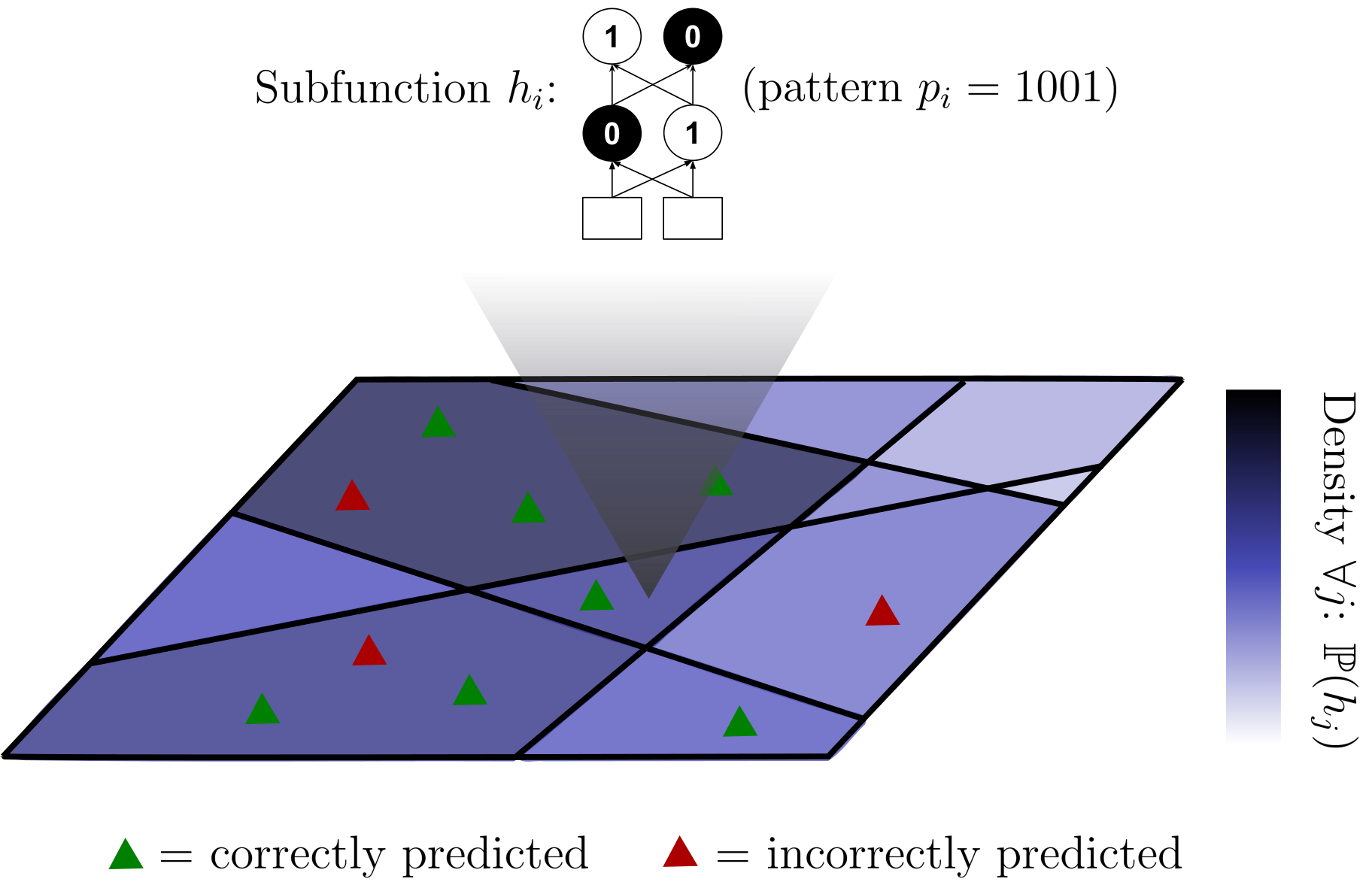}
 \caption{Each linear activation region in 2D input space (plane) is mapped to a unique subfunction, activation decision pattern, and set of training samples (triangles). A density, smooth in representation space, is defined given the number of samples in each region.}
 \label{f:notation}
\end{wrapfigure}

This is a Hoeffding inequality based bound \citep[][eq. 2.17]{mohri2018foundations}.
As we take a pre-trained model and rank test samples, the model is fixed.
%; what varies, and is marginalized out by expectation, is the draw of test data.
There are several drawbacks with this initial formulation. 
First, it treats the empirical error of different subfunctions as independent 
when in general, they are not. Since different activation regions are bounded by shared hyperplanes, and hence the subfunctions share parameters, there exists useful evidence for a subfunction's performance outside its own activation region. Second, test samples overwhelmingly induce unseen activation patterns in deep networks, and the bound in \cref{eq:hyperlocal_bound} is infinite if $N_i = 0$, making this quantity uninformative for the purposes of comparing subfunctions.
%out-of-sample. 

This motivates the following empirical risk metric for subfunctions. 
Fix non-negative weighting or closeness distance function between subfunctions, $k: H \times H \rightarrow \mathcal{R}^{\geq 0}$.
For any subfunction $h_i \in H$, define its probability mass:
\begin{align}
    & \prob(h_i) \eqdef 
    \frac{\sum_{j \in [1, C]} N_j \spa k(h_i, h_j) }{ \sum_{l \in [1, C]} \sum_{j \in [1, C]} N_j  \spa k(h_l, h_j)} \label{eq:density}, \\ 
    & \text{so by construction} \sum_{i \in [1, C]} \prob(h_i) = 1,
\end{align}
which quantifies how densely $h_i$ is locally populated by training samples.
%, local meaning in subfunction kernel space, not Euclidean closeness in input space; though note these are related as Euclidean distance of 0 between samples implies the same induced subfunction.
Rewrite the empirical error of the network as an expectation over subfunction empirical error:
\begin{align}
    \emprisk_S(f) =& \frac{1}{N} \sum_{(x, y) \in S} r((x, y); f) \label{eq:dist_weighted_fact0} \\
    =& \frac{1}{N} \sum_{(x, y) \in S} \frac{1}{\sum_{j \in [1, C]} \prob(h_j) \spa k(H(x), h_j)} \sum_{i \in [1, C]} \prob(h_i) \spa k(H(x), h_i) \spa r((x, y); f) \label{eq:dist_weighted_fact1} \\
    =& \frac{1}{N} \sum_{i \in [1, C]} \prob(h_i) \sum_{(x, y) \in S} 
    \frac{k(H(x), h_i) \spa r((x, y); f)}{ \sum_{j \in [1, C]} \prob(h_j) \spa k(H(x), h_j) } \label{eq:dist_weighted_fact2} \\
    =& \ex_{h_i} \bigg [ \frac{1}{N} \sum_{(x, y) \in S} \frac{k(H(x), h_i) \spa r((x, y); f)}{ \sum_{j \in [1, C]} \prob(h_j) \spa k(H(x), h_j)  } \bigg ] \label{eq:dist_weighted_fact3} \\
    =& \ex_{h_i} \bigg [ \frac{1}{N} \sum_{l \in [1, C]} \frac{k(h_l, h_i) \spa N_l \spa \emprisk_{S_{l}}(h_l)}{ \sum_{j \in [1, C]} \prob(h_j) \spa k(h_l, h_j) } \bigg ] \label{eq:dist_weighted_fact4} \eqdef \ex_{h_i} [ \emprisk^*_{S} (h_i) ]. %\label{eq:dist_weighted_fact5}
\end{align}

%\newpage
\begin{theorem}[Expected smooth error bound]\label{thm:true_error}
Assume that $\tilde{D}_S$ is a distribution over size-$N$ datasets that have the same activation region data distributions and dataset sizes ($D_i$ and $N_i$, $\forall i \in [1, C]$) as $S$. Let dataset $S$ be drawn iid from $\tilde{D}_S$.
Then $\forall i \in [1, C]$, for any given $\delta \in (0, 1]$, with probability $> 1 - \delta$:
\begin{align}
    \risk^*_{\tilde{D}_S}(h_i) \eqdef \ex_S [\emprisk^*_{S}(h_i)] \leq \emprisk^*_{S}(h_i) + \frac{1}{\prob(h_i)} \sqrt{\frac{\log \frac{2}{\delta}}{2 N}}, \label{eq:true_error_bound}
\end{align}
\end{theorem}

\noindent proof in \cref{app:1}.
%via a variant of Hoeffding's inequality (\cref{app:1}).
Intuitively this implies that the more a subfunction is surrounded by samples for which the model makes accurate predictions - both from its own region and regions of other subfunctions, weighted by the weighting function $k$ -  the lower its empirical error and bound on generalization gap, and thus the lower its bound on true or expected error, given any $\delta$.  
The further that test samples fall from densely supported training regions or subfunctions, the less likely the model is to be accurate.
Unlike \cref{eq:hyperlocal_bound}, this bound is finite even for subfunctions without training samples because such subfunctions are assigned non-zero density (\cref{f:notation}), given a positive-valued weighting function. 

Smoothing is performed out of necessity; in order to resolve the problem of not having empirical training samples to quantify the error of a subfunction, one must assume interdependence between subfunctions.
%as empirical quantities cannot be measured without empirical samples.
In general, smoothing with $k$ introduces a bias since the bound on a subfunction's smooth empirical error does not converge to expected error in the limit of number of samples of its region. 
However, smoothing is not unreasonable as errors of different subfunctions in neural networks are interdependent due to parameter sharing, and the bias is reducible by searching $k$, which experimental results indicate is sufficient to render it inconsequential in practice.

\subsection{Weighting function}\label{s:weighting_function}

The bound in \cref{thm:true_error} holds irrespective of choice of $k$ because the bounding procedure assumes a worst case. 
This makes the bound loose, but allows for $k$ to be searched using a validation set. 
In this section we discuss selecting $k$.
%Although $\frac{1}{\prob(h_i)}$ can be very large in general, note firstly that it is tempered by a factor of $\frac{1}{\sqrt{N}}$, and secondly that if one were to optimize the weighting function $k$ by minimizing the distance between true error and true smooth error (\cref{eq:true_error_bound}), this necessarily involves finding a function that yields high densities for activation regions, otherwise the distance to true error would be large.
%
%Using the bound in \cref{eq:true_error_bound} for practical purposes requires providing weighting function $k$, as well as probability $\delta$. 
The ideal weighting function $k$ and probability parameter $\delta$ produce a bound for each subfunction $h_i$ that
%where the empirical risk
reflects the subfunction's true error accurately, i.e. minimizes the difference with the expected true error (without weighting function) if we had unlimited samples of its activation region $a_i$:
%and 2) is tight, i.e. with small generalization gap:

\begin{align}
    \min_{k, \delta} \spa 
    \Bigg \lVert \ex_{S_i} [\emprisk_{S_i}(h_i)]  - \Bigg (  \emprisk^*_{S}(h_i) 
    + \frac{1}{\prob(h_i)} \sqrt{\frac{\log \frac{2}{\delta}}{2 N}} \Bigg ) \Bigg \rVert 
    \label{eq:weight_obj}
\end{align}

Being limited to dataset instance $S$, we do not have access to subfunction $h_i$'s true error $\ex_{S_i} [\emprisk_{S_i}(h_i)]$. 
However, we can construct an estimate by taking the samples $x, y$ 
that $k$ includes in the subfunction's smooth empirical error $\emprisk^*_{S}(h_i)$ (that is, all the training samples, for positive valued $k$) and adjusting their error values to reflect what they would be if the sample did belong to $a_i$. This improves on $\emprisk^*_{S}(h_i)$, the weighted error of surrounding activation regions, by transforming it into a weighted error of the target activation region $a_i$ itself.

\begin{wrapfigure}[16]{l}{0.33\textwidth}
 \centering
 \includegraphics[width=0.33\textwidth]{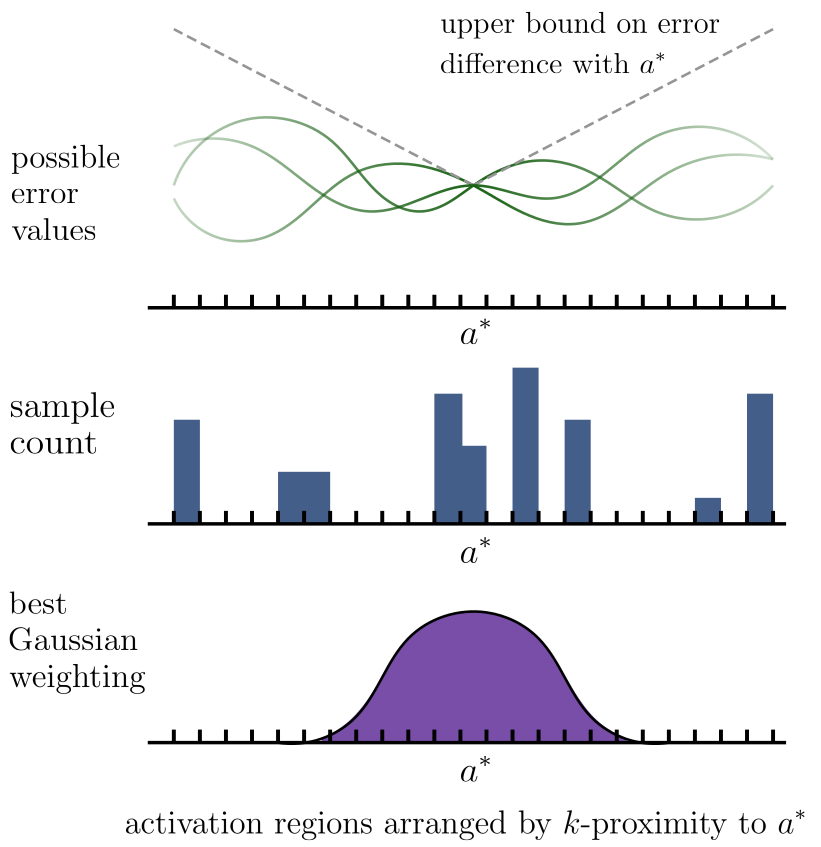}
 \caption{
 Given a family of Gaussian weighting functions, 
 the best weighting function according to \cref{eq:weighting_obj_final} trades off precision of the smooth empirical error with sample density.}
 \label{f:weighting}
\end{wrapfigure}

Let $\Phi: \mathcal{X} \rightarrow  \mathcal{Z}$ be any data representation for which there exists a function $\Psi: \mathcal{Z} \rightarrow \reals$ that computes inference model $f$'s per-sample error, i.e. $r(x, y) = \Psi(\Phi(x)) $ for all $x, y$ in the support of $D$. This assumes the data is well conditioned so target labels are predictable from inputs.
Let $\Omega: \mathcal{Z} \rightarrow \mathcal{Z}$ be a shifting function that changes the representation of any sample $x \in \mathcal{X}$ into the representation of a sample in $a_i$, $\Omega(\Phi(x)) \eqdef \argmin_{\Phi(x')} \lVert \Phi(x') - \Phi(x) \rVert \text{ s.t. } x' \in a_i$.
%Then difference in error scales with O|phi(x) - Omega(phi(x'))|
%or, include y separately?
%does it hold if Z is neural network repr?
%therefore, should upweight close representational distance
%
%Let $g : \mathcal{X} \times \mathcal{Y} \rightarrow \mathcal{X} \times \mathcal{Y} $ be the shifting function that moves samples into region $a_i$, defined as $g(x, y) \eqdef \argmin_{(x', y') \sim D} \lVert (x', y') - (x, y) \rVert \text{ s.t. } x' \in a_i$.
%
Using the Taylor expansion, assuming an error function $\Psi$ with bounded first order gradients:
\begin{align}
    \forall (x, y) \in S: \Psi(\Omega(\Phi(x)))= \Psi(\Phi(x)) + \mathcal{O}(\lVert \Phi(x) - \Omega(\Phi(x)) \rVert)
\end{align}

Replacing the true error in \cref{eq:weight_obj} with the estimate constructed using shifted samples:
\begin{align}
  %\min_{k, \delta} \spa 
  %
  &\Bigg \lVert \frac{1}{N} \sum_{(x, y) \in S} \frac{1}{w(x)} k(H(x), h_i) \spa \Psi(\Omega(\Phi(x)))
  -
  \Bigg (  \emprisk^*_{S}(h_i)
  + \frac{1}{\prob(h_i)} \sqrt{\frac{\log \frac{2}{\delta}}{2 N}}  \Bigg )  \Bigg \rVert \\
  %
 % = \min_{k, \delta} \spa 
   \leq& \Bigg \lVert \frac{1}{N} \sum_{(x, y) \in S} \frac{1}{w(x)} k(H(x), h_i) \spa \mathcal{O}(\lVert \Phi(x) - \Omega(\Phi(x)) \rVert)  \Bigg \rVert
  + 
   \frac{1}{\prob(h_i)} \sqrt{\frac{\log \frac{2}{\delta}}{2 N}}, \label{eq:weighting_obj_final}
\end{align}

\vspace{0.25em}

where $w(x) \eqdef \sum_{j \in [1, C]} \prob(h_j) \spa k(H(x), h_j)$ is the weight normalization term.
We draw 2 conclusions from this analysis. First, weighting value $k(h_i, h_j)$ for any $h_i, h_j$ should decrease with the distance between the feature representations of samples in their activation regions. This is evident from \cref{eq:weighting_obj_final} as $\Phi(x)$ is the representation of a sample in the activation region of $H(x)$, and $\Omega(\Phi(x))$ is the representation of a sample in the activation region of $h_i$, so larger distances should be penalized by smaller weights. However, $k(h_i, h_j)$ cannot be too low (for example 0 for all $h_j \neq h_i$ given any $h_i$) because the magnitude of $\frac{1}{\prob(h_i)}$ would be large. Thus the best $k$ combines error precision (upweight near regions and downweight far regions) with support (upweight near and far regions). The need to minimize weight with distance justifies restricting the search for $k$ within function classes where output decreases with representation distance, such as Gaussian functions of activation pattern distance.
An example of this trade-off is illustrated in \cref{f:weighting}. 

Secondly, in practice there is no need to explicitly find $k$ that minimizes \cref{eq:weighting_obj_final}. A simple alternative is to use a validation set metric that correlates with how well $k, \delta$ minimize \cref{eq:weighting_obj_final} (i.e. approximate the true error), such as the ability of the bound to discriminate between validation samples that are accurately or inaccurately predicted, and select $k$ and $\delta$ such that they minimize the validation metric. This is the approach we take in our experiments.

\vspace{-0.5em}
\section{Related work}
\vspace{-0.5em}
This work is primarily related to sample-level metrics intended for out-of-distribution or unreliable in-distribution sample selection \citep{ovadia2019can}, and work on linear activation regions, which is typically motivated by characterizing neural network expressivity \citep{montufar2014number,raghu2017expressive,hanin2019deep}.

Well known sample uncertainty or unreliability metrics include the maximum response of the final softmax prediction layer  \citep{hendrycks2016baseline,geifman2017selective,cordella1995method,chow1957optimum}, its entropy \citep{shannon1948mathematical}, or its top-two margin \citep{scheffer2001active}, all conditioned on the input sample.
\cite{liang2017enhancing} combines maximum response with temperature scaling and input perturbations.
\cite{jiang2018trust} combines the top-two margin idea with class distance.
Some ideas use distance to prototypes in representation space, which is similar at high level to ours if one assumes prototypes are in high-density regions.
\cite{lee2018simple} trains a logistic regressor on layer-wise distances of a sample's features to its nearest class, with the idea that distance to features of the nearest class should scale with unreliability. This was shown to outperform \cite{liang2017enhancing}. \cite{sehwag2021ssd} is an unsupervised variant of \cite{lee2018simple}.
\cite{tack2020csi} clusters feature representations instead of using classes, using distance to nearest cluster as unreliability.
In \citet{bergman2020classification} the clusters are defined by input transformations; we were unable to get this working in our setting as models appear to suffer from feature collapse across transformations when not trained explicitly for transformation disentanglement.
%Note that our work is specifically about determining what a learned model does not know, rather than OOD for its own sake. 
%often only care about outliers in the context of a predictive model for an actual task i.e. using information from that model itself. Note also that in many practical scenarios one does not have the resources or inclination to train separate neural networks specifically for the task of OOD detection, in particular as this would not catch the in-distribution mispredictions. Theoretically, using the predictive model also makes the problem interesting because it’s about exposing what a predictive model does not know using the model itself.
Non-parametric kernel based methods such as Gaussian processes provide measures of uncertainty that also scale with distance from samples and can be appended to a neural network base \citep{liu2020simple}.
\cite{zhang2020hybrid} assume density in latent space 
is correlated to reliability, using residual flows \citep{chen2019residual} for the density model.
If multiple models trained on the same dataset are available (which we do not assume), one could use ensemble model metrics such as variance, max response or entropy \citep{lakshminarayanan2016simple}; an ensemble can also be simulated in a single model with Monte Carlo dropout 
\citep{geifman2017selective, gal2016dropout}.

%To set this work in context, we cast generalization error bounds as a sample-level uncertainty metric, by dividing the network into activation regions whose functions are co-dependent.
%Our sample-level experiments do not deal with training neural networks to maximize generalization (model selection), but analyze trained models.
%By uncertainty we mean epistemic uncertainty (reducible by increasing sample size) as opposed to aleatoric (e.g. measurement error).
Many of these works seek to predict whether a sample is out-of-distribution (OOD) for its own sake, which is an interesting problem, but we care about 1) epistemic uncertainty in general, including in-distribution misclassification, not just OOD 2) in the context of the main model trained for a practical task, or in other words, exposing what the task model does not know using the task model itself, as opposed to training separate models on the data distribution specifically for outlier detection.

\vspace{-0.5em}
\section{Experiments}
\vspace{-0.5em}

We tested the ability of the input-conditioned bound in \cref{eq:true_error_bound} to predict out-of-distribution and misclassified in-distribution samples.
%for image classification tasks compared to a variety of baseline methods. 
Taking pre-trained VGG16 and ResNet50 models for CIFAR100 and CIFAR10, we computed area under false positive rate vs. true positive rate (AUROC) and area under coverage vs. effective accuracy (AUCEA) for each method. Definitions are given in \cref{app:metrics}. These metrics treat predicting unreliable samples as a binary classification problem, where for out-of-distribution, ground truth is old distribution/new distribution, and for misclassified in-distribution, ground truth is classified correct/incorrect. Method output is accept/reject. All methods produce a metric per sample that is assumed to scale with unreliability, so 1K thresholds for discretizing into accept/reject decisions were uniformly sampled across the maximum test set range, yielding the AUROC and AUCEA curves. 
For our method, we used Gaussian weighting with standard deviation $\rho$, $k(h_i, h_j) = e^{- \operatorname{Hamming}(p_i,p_j)^2 / (2\rho^2)}$, and took log of the bound to suppress large magnitudes.
In-distribution misclassification validation set was used to select all hyperparameters, i.e. we use a realistic, hard setting where OOD data is truly unseen for all parameters. 

% averages table
\begin{table}[b]
\vspace{-0.5em}
\centering
\footnotesize
\setlength{\belowcaptionskip}{-0.5em}
\caption{Summary of out-of-distribution and misclassified in-distribution results, by difference to the top performing method in each architecture $\times$ dataset setting. Values are difference in AUROC and average $\pm$ standard deviation is shown over all architecture $\times$ dataset settings. Higher is better.}\label{t:avg}
\begin{tabular}{l c c c}
\toprule
& Out-of-distr. & Misc. in-distr. & Average \\
\midrule
Residual flows density \citep{chen2019residual} & -0.538 $\pm$ 2E-01  & -0.356 $\pm$ 4E-02  & -0.447 $\pm$ 1E-01 \\
GP (\cite{liu2020simple} w/ fixed features) & -0.204 $\pm$ 2E-01  & -0.159 $\pm$ 1E-01  & -0.181 $\pm$ 1E-01 \\
Class distance \citep{lee2018simple} & -0.214 $\pm$ 1E-01  & -0.334 $\pm$ 9E-02  & -0.274 $\pm$ 1E-01 \\
Margin \citep{scheffer2001active} & -0.037 $\pm$ 2E-02  & -0.007 $\pm$ 7E-03  & -0.022 $\pm$ 1E-02 \\
Entropy \citep{shannon1948mathematical} & -0.025 $\pm$ 2E-02  & \textbf{-0.002 $\pm$ 2E-03}  & -0.014 $\pm$ 1E-02 \\
Max response \citep{cordella1995method} & -0.034 $\pm$ 2E-02  & -0.008 $\pm$ 8E-03  & -0.021 $\pm$ 1E-02 \\
MC dropout \citep{geifman2017selective} & -0.061 $\pm$ 3E-02  & -0.048 $\pm$ 2E-02  & -0.054 $\pm$ 2E-02 \\
Cluster distance \citep{tack2020csi} & -0.052 $\pm$ 9E-02  & -0.021 $\pm$ 7E-03  & -0.036 $\pm$ 5E-02 \\
Subfunctions (ours) & \textbf{-0.007 $\pm$ 1E-02}  & -0.006 $\pm$ 4E-04  & \textbf{-0.007 $\pm$ 6E-03} \\
\bottomrule
\end{tabular}
%\vspace{-0.3em}
\end{table}

We note 2 adjustments to the theory made in the practical experiments. First, for tractability on deep networks, we use a coarse partitioning of the network by taking activations from the last ReLU layer only, so the subfunctions are piecewise linear instead of purely linear. In this case activation regions are still disjoint, \cref{eq:true_error_bound} becomes a bound on piecewise linear subfunction error and still holds. Second, computing the set of feasible activation regions is intractable, so we use the full bit space, $P = \{0, 1\}^{M}$. This means the bound is computed in an altered subfunction space where some infeasible subfunctions are assigned a non-zero density, affecting the weight normalization. A benefit of using the full bit space is it allows computational savings when computing the bound, detailed in \cref{s:practical_derivations}. 
These 2 limitations mean that the performance attained by the method in the experiments is a lower bound that is likely improvable if more tractable implementations are found.

\begin{wrapfigure}[27]{l}{0.48\textwidth}
%\vspace{-1.25em}
 \centering
 \includegraphics[width=0.48\textwidth]{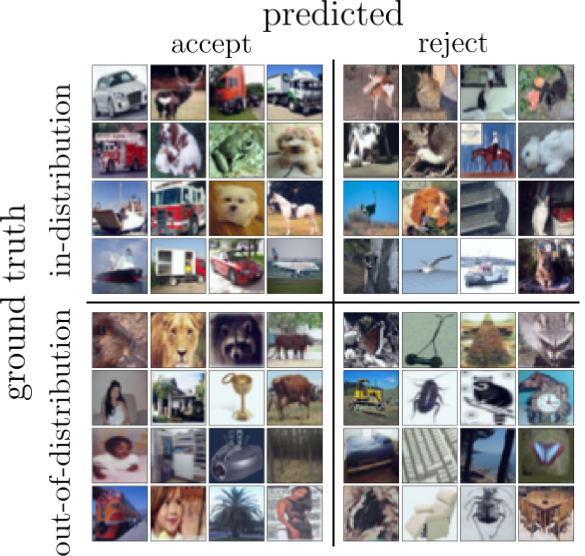}
 \setlength{\abovecaptionskip}{-0.5em}
 \caption{Sample confusion matrix, OOD for CIFAR10 $\rightarrow$ CIFAR100 on ResNet50. Random samples from top 20\% in each quadrant shown.}
 %\vspace{-1em}
 \label{fig:qual}
\end{wrapfigure}

Subfunctions and entropy were found to be the top 2 methods overall in each category 
(\cref{t:avg}), with subfunctions better on average. Cluster distance also performed well on CIFAR10, but was penalized by poor performance on CIFAR100 and particularly VGG16 (\cref{t:ood_cifar100,t:id_cifar100}), which is a more difficult dataset for determining outlier status as the in-distribution classes are more finegrained.
We conclude that subfunctions and entropy were good predictors of unreliability for both in-distribution and out-of-distribution scenarios. The AUCEA results for the in-distribution setting (\cref{t:id_cifar10,t:id_cifar100}) mean that using either to filter predictions would have raised effective model accuracy (accuracy of accepted samples) to $90\sim 92\%$ from $70\sim73\%$ for the original CIFAR100 models,
and to $98\sim99\%$ from $91\sim92\%$ for the original CIFAR10 models (\cref{t:orig_acc}), on average over thresholds.
Entropy is simpler and computationally cheaper than subfunction error bound, but suffers from the drawback that it can only be computed exactly if model inference includes a discrete probabilistic variable. This is the case for these experiments but not in general 
(e.g. consider MSE objectives), whereas our method does not have this restriction.

\begin{table}
    \centering
    \footnotesize
    \caption{Results for models trained on CIFAR10 on out-of-distribution detection vs CIFAR100/SVHN. AUROC shown, higher is better. For equivalent table on CIFAR100, see \cref{t:ood_cifar100}.}
    \label{t:ood_cifar10}
    \begin{tabular}{l c c c c}
\toprule
 & \multicolumn{2}{c}{$\rightarrow$ CIFAR100}  & \multicolumn{2}{c}{$\rightarrow$ SVHN} \\
\midrule
 & VGG16  & ResNet50  & VGG16  & ResNet50 \\
\midrule
Residual flows density & 0.513 $\pm$ 2E-04  & 0.513 $\pm$ 1E-04  & 0.084 $\pm$ 1E-04  & 0.084 $\pm$ 1E-04 \\
GP & 0.810 $\pm$ 1E-02  & 0.575 $\pm$ 2E-02  & 0.844 $\pm$ 2E-02  & 0.473 $\pm$ 9E-02 \\
Class distance & 0.673 $\pm$ 7E-02  & 0.468 $\pm$ 3E-02  & 0.806 $\pm$ 6E-02  & 0.462 $\pm$ 1E-01 \\
Margin & 0.829 $\pm$ 2E-03  & 0.825 $\pm$ 5E-03  & 0.854 $\pm$ 4E-02  & 0.856 $\pm$ 2E-02 \\
Entropy & 0.853 $\pm$ 2E-03  & 0.822 $\pm$ 6E-03  & 0.869 $\pm$ 3E-02  & 0.858 $\pm$ 3E-02 \\
Max response & 0.829 $\pm$ 3E-03  & 0.827 $\pm$ 5E-03  & 0.850 $\pm$ 4E-02  & 0.858 $\pm$ 2E-02 \\
MC dropout & 0.776 $\pm$ 6E-03  & 0.807 $\pm$ 3E-03  & 0.778 $\pm$ 6E-02  & 0.838 $\pm$ 3E-02 \\
Cluster distance &  \textbf{0.862 $\pm$ 4E-03 }  &  \textbf{0.867 $\pm$ 3E-03 }  &  \textbf{0.870 $\pm$ 5E-02 }  &  \textbf{0.892 $\pm$ 1E-02 } \\
Subfunctions (ours) &  \textbf{0.858 $\pm$ 4E-03 }  &  \textbf{0.862 $\pm$ 2E-03 }  &  \textbf{0.886 $\pm$ 2E-02 }  &  \textbf{0.915 $\pm$ 2E-02 } \\
\bottomrule
\end{tabular}
%\vspace{-1.5em}
\end{table}

\begin{table}
    \centering
    \footnotesize
    %\vspace{-0.5em}
    %\setlength{\belowcaptionskip}{-0.5em}
    \caption{Results for models trained on CIFAR10. Predicting misclassification on in-distribution test. Higher is better. For equivalent table on CIFAR100, see \cref{t:id_cifar100}.}
    \label{t:id_cifar10}
    \begin{tabular}{l c c c c}
\toprule
 & \multicolumn{2}{c}{VGG16}  & \multicolumn{2}{c}{ResNet50} \\
\midrule
 & AUCEA  & AUROC  & AUCEA  & AUROC \\
\midrule
Residual flows density & 0.442 $\pm$ 5E-02  & 0.520 $\pm$ 1E-02  & 0.492 $\pm$ 3E-02  & 0.577 $\pm$ 1E-02 \\
GP & 0.983 $\pm$ 1E-03  & 0.865 $\pm$ 8E-03  & 0.943 $\pm$ 5E-03  & 0.625 $\pm$ 2E-02 \\
Class distance & 0.948 $\pm$ 2E-02  & 0.669 $\pm$ 8E-02  & 0.900 $\pm$ 3E-02  & 0.471 $\pm$ 6E-02 \\
Margin & 0.982 $\pm$ 2E-03  & 0.900 $\pm$ 4E-03  & 0.848 $\pm$ 1E-02  &  \textbf{0.894 $\pm$ 4E-03 } \\
Entropy &  \textbf{0.989 $\pm$ 1E-04 }  &  \textbf{0.914 $\pm$ 3E-03 }  &  \textbf{0.984 $\pm$ 1E-03 }  & 0.890 $\pm$ 5E-03 \\
Max response & 0.980 $\pm$ 3E-03  & 0.898 $\pm$ 4E-03  & 0.832 $\pm$ 1E-02  &  \textbf{0.895 $\pm$ 5E-03 } \\
MC dropout & 0.982 $\pm$ 6E-04  & 0.845 $\pm$ 6E-03  & 0.976 $\pm$ 2E-03  & 0.868 $\pm$ 1E-02 \\
Cluster distance & 0.988 $\pm$ 4E-04  & 0.901 $\pm$ 2E-03  & 0.981 $\pm$ 2E-03  & 0.867 $\pm$ 2E-03 \\
Subfunctions (ours) &  \textbf{0.988 $\pm$ 3E-04 }  &  \textbf{0.907 $\pm$ 3E-03 }  &  \textbf{0.983 $\pm$ 2E-03 }  & 0.889 $\pm$ 6E-03 \\
\bottomrule
\end{tabular}
\end{table}

\begin{figure}[h]
    \centering
    \includegraphics[width=\textwidth]{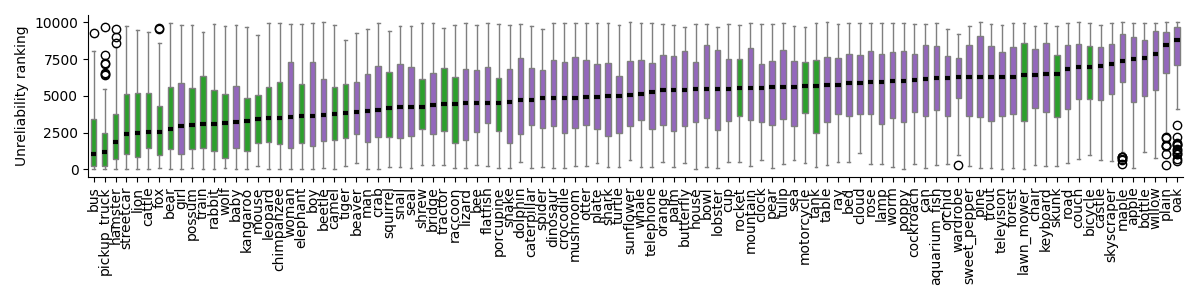}
    \setlength{\abovecaptionskip}{-1em}
    \setlength{\belowcaptionskip}{-1em}
    \caption{OOD for CIFAR10 $\rightarrow$ CIFAR100 on ResNet50. 10k CIFAR100 test samples were ranked by unreliability (log STEB). Boxplots summarize rankings per class (lower = less unreliable). Green denotes superclasses similar to CIFAR10: carnivores, omnivores, herbivores, mammals, vehicles. 
    %Similar CIFAR100 classes appear as more reliable to the CIFAR10 model, according to STEB.
    }
    \label{fig:boxplot}
\end{figure}

Empirically, we observed for subfunctions that reliable in-distribution images tended to be prototypical images for their class, whilst OOD images erroneously characterized as 
reliable tended to resemble the in-distribution classes, as one would expect (\cref{fig:qual}). 
%This figure contains random samples from the $20\%$ most reliable/unreliable samples of in-distribution and OOD test sets, corresponding to the accept/reject columns.
To test this hypothesis further, we took a CIFAR10 model and plotted the rankings of samples from each CIFAR100 class in \cref{fig:boxplot}, where each box denotes the median and first and third quartiles.
%, whiskers have length 1.5 * interquartile range, and circles denote outliers outside of whisker range. 
The classes are ordered by median. 
We identified the superclasses in CIFAR100 with the highest semantic overlap with CIFAR10 classes (made up of mostly vehicles and mammals), whose classes are coloured green. It is clear from the correlation between green and lower unreliability ranking that subfunction error bounds rate OOD classes semantically closer to the training classes as more reliable. Note that even the exceptions towards the right are justified, because the inclusion of e.g. bicycles, lawn mowers and rockets in ``vehicles'' is questionable; certainly these objects do not correspond visually to the vehicle classes in CIFAR10. In addition, we ran the same plot for the other methods and found that in every case the correlation between reliability and semantic familiarity was less strong (\cref{app:boxplots}).

%Note that our method generalizes well to detecting out-of-distribution samples taken from datasets different to the training dataset, even though the derivation of the bound assumes that test and training distributions are identical. 
%One justification is that OOD data can be viewed as an unlucky draw of in-distribution data, as long as it is included in the support of the training distribution, which is reasonable in the case of natural images. 
%

\section{Conclusion}

Density of training samples in representation space appears to be a feasible indicator of reliability of predictions for trained piecewise linear neural networks. This raises several interesting questions for future work:

\begin{itemize}
    \item Measures of unreliability that scale with density of continuous input samples in representation space, rather than density of discrete partitions, which is specialized to piecewise linear neural networks. 
    \item Deriving tighter bounds, for example by making stronger assumptions about the weighting function $k$ used.
    \item Implications for model selection; how to train networks such that samples fall in high-density regions in representation space.  
\end{itemize}

With regards to model selection, a reasonable hypothesis based on our results is that generalization ability of neural networks scales with the proportion of test inputs mapped to high-density regions in its representation space. Low-density activation regions are less likely in compact representation spaces, all else equal, since the same number of training samples is distributed over fewer representations.
This is an intuition rather than a formal result of this work, but it links to a body of work on the relation between compact representation spaces and generalization.
%maximising the abstractness of the model's data representation. 
Compactness is optimized for in information bottlenecks \citep{tishby2000information,ahuja2021invariance}, which minimize the entropy of network representations, and implicitly by sparse factor graphs \citep{goyal2020inductive} and feature discretization methods \citep{dianbo_liu2021discrete,van2017neural} including discrete output unsupervised learning \citep{ji2019invariant}, which are methods that build low expressivity into the model as a prior for improved generalization. 
We conclude that learning compact representation spaces with few, densely supported modes is an interesting direction for future work on neural network generalization. 

\bibliography{collas2022_conference}
\bibliographystyle{collas2022_conference}

\newpage
\appendix 

%\begin{center}
%    {\Large{\textbf{Predicting Unreliable Predictions by Shattering a Neural Network: Supplementary Material}}}
%\end{center}

% \section{Algorithm} \label{s:algo}
% \input{algo}

\section{Proof for \Cref{thm:true_error}}\label{app:1}
This follows the Chernoff/Hoeffding bounding technique.
Fix model $f$, patterns $P$ and thus subfunctions set $H$.
Let the true data distribution for each activation region $a_i$ be $D_i$.
%so $D_i$ has zero probability mass for any input sample $x \notin a_i$.
Assume that $\tilde{D}_S$ is a distribution over size-$N$ datasets that have the same activation region data distributions and dataset sizes ($D_i$ and $N_i$, $\forall i \in [1, C]$) as $S$. Consider $S$ as a dataset drawn iid from $\tilde{D}_S$.
Without loss of generality, since dataset order is immaterial for the error metrics, assume the index $n$ of sample $S^n$ determines its activation region and distribution from which it is drawn independently of other samples.
That is, 
$\forall n \in [1, N]: S^n \sim D_{J^n} \text{ iid}$ for some fixed $J \in [1, C]^N$.
Recall that $D_i$ is constructed as $\prob_{D_i}(x, y) = \prob_D(x, y | x \in a_i)$.
Call the inputs and targets $X_S$ and $Y_S$, $\forall n \in [1, N]: (X_S^n, Y_S^n) \eqdef S^n$.
Recall that empirical error function for samples is bounded: $\forall x, y \in (\mathcal{X} \times \mathcal{Y}): 0 \le r((x, y); f) \le 1$ and weighting function $k$ is non-negative valued.

Choose any subfunction $h_i \in H$, $i \in [1, C]$.
%As with \citet[][eq. 2.17]{mohri2018foundations},we bound the subfunction's true error using Hoeffding's inequality \citep{hoeffding1994probability}. 
Then $\forall \epsilon > 0, t > 0, \tilde{S} \sim \tilde{D}_S$ iid:
\begin{align}
    & \prob_{S} (\emprisk^*_{S}(h_i) - \risk^*_{\tilde{D}_S}(h_i) \ge \epsilon ) = 
    \prob_{S} (\emprisk^*_{S}(h_i) - \ex_{\tilde{S}} [\emprisk^*_{\tilde{S}}(h_i)] \ge \epsilon ) \label{eq:dist_weighted_0} \\
    &= \prob_{S} (e^{t(\emprisk^*_{S}(h_i) - \ex_{\tilde{S}} [\emprisk^*_{\tilde{S}}(h_i)])} \ge e^{t\epsilon}) \label{eq:dist_weighted_1} \\
    &\le e^{-t\epsilon} \ex_{S} [ e^{t(\emprisk^*_{S}(h_i) - \ex_{\tilde{S}} [\emprisk^*_{\tilde{S}}(h_i)])} ] 
    %\text{\hspace{15em}(Markov's inequality)}  
    \label{eq:dist_weighted_2} \\
    &= e^{-t\epsilon} \ex_{S} [ e^{t \sum_{n=1}^N
    \big ( \frac{k(H(X_S^n), h_i) \spa r(S^n; f)}{ N \sum_{j \in [1, C]} \prob(h_j) \spa k(H(X_S^n), h_j) } - 
    \ex_{\tilde{S}^n} 
    \big[ 
    \frac{k(H(X_{\tilde{S}}^n), h_i) \spa r(\tilde{S}^n; f)}{ N \sum_{j \in [1, C]} \prob(h_j) \spa k(H(X_{\tilde{S}}^n), h_j) }
    \big] \big ) } ] \label{eq:dist_weighted_3} \\
    &= e^{-t\epsilon} \prod_{n = 1}^{N} \ex_{S^n} [ e^{t
    \big ( \frac{k(H(X_S^n), h_i) \spa r(S^n; f)}{ N \sum_{j \in [1, C]} \prob(h_j) \spa k(H(X_S^n), h_j) } - 
    \ex_{\tilde{S}^n} 
    \big[ 
    \frac{k(H(X_{\tilde{S}}^n), h_i) \spa r(\tilde{S}^n; f)}{ N \sum_{j \in [1, C]} \prob(h_j) \spa k(H(X_{\tilde{S}}^n), h_j) }
    \big] \big ) } ] \label{eq:dist_weighted_4} \\
    &\le e^{-t\epsilon} \prod_{n = 1}^{N} e^{\frac{ t^2 \big ( \frac{1}{N \prob(h_i)} - 0 \big )^2}{8}}
    %\text{\hspace{17em} (Hoeffding's lemma)}
    \label{eq:dist_weighted_5} \\
    &= e^{\frac{t^2}{8 N \prob(h_i)^2} - t\epsilon} \label{eq:dist_weighted_6} ,
\end{align}
by making use of the following:
\begin{enumerate}
    \item \cref{eq:dist_weighted_1} $\rightarrow$ \cref{eq:dist_weighted_2} : Markov's inequality. For random variable $Z \ge 0$ and constant $a > 0$, $\prob (Z \ge a) \leq \frac{\ex [Z]}{a}$.
    
    \item \cref{eq:dist_weighted_2} $\rightarrow$ \cref{eq:dist_weighted_3} : definition of $\emprisk^*_{S}(h_i)$ and linearity of expectation.
    
    \item \cref{eq:dist_weighted_3} $\rightarrow$ \cref{eq:dist_weighted_4} : $\emprisk^*_{S}(h_i)$ is a sum over independently drawn samples. $\forall h_j \in H: \prob(h_j)$ is constant because activation region dataset sizes are constant.
    
    \item \cref{eq:dist_weighted_4} $\rightarrow$ \cref{eq:dist_weighted_5} : first bound the length of the range of the exponent. 
    $\forall i \in [1, C], n \in [1, N]: 0 \le \frac{k(H(X_S^n), h_i) \spa r(S^n; f)}{ N \sum_{j \in [1, C]} \prob(h_j) \spa k(H(X_S^n), h_j) } 
    \le 
    \frac{k(H(X_S^n), h_i)}{ N \prob(h_i) \spa k(H(X_S^n), h_i) } = \frac{1}{N \prob(h_i)}$.
    Subtracting a constant does not change the length of the range of a random variable.
    Then apply Hoeffding's lemma: for random variable $Z$ where $ a \le Z \le b$ and $\ex [Z] = 0$, then $\forall t > 0: \ex [e^{tZ}] \le e^{\frac{t^2(b - a)^2}{8}}$ holds. % , b > a
\end{enumerate}
Find the optimal $t$ as the one that yields the tightest bound:
\begin{align}
    \nabla_t e^{\frac{t^2}{8N \prob(h_i)^2} - t\epsilon}  = 0, \qquad t = 4 N \epsilon \prob(h_i)^2 \label{eq:dist_weighted_7}, 
\end{align}
which is a minimum because the second derivative is positive. Substitute into \cref{eq:dist_weighted_6} :
\begin{align}
    \prob_{S} (\emprisk^*_{S}(h_i) - \risk^*_{\tilde{D}_S}(h_i) \ge \epsilon ) \le e^{-2 N \epsilon^2 \prob(h_i)^2} \label{eq:dist_weighted_8} .
\end{align}
The symmetric case can be proved in the same way:
\begin{align}
    \prob_{S} (\emprisk^*_{S}(h_i) - \risk^*_{\tilde{D}_S}(h_i) \le -\epsilon ) = \prob_{S} (\risk^*_{\tilde{D}_S}(h_i) - \emprisk^*_{S}(h_i)  \ge \epsilon ) \le e^{-2 N \epsilon^2 \prob(h_i)^2} \label{eq:dist_weighted_9} ,
\end{align}
\noindent specifically because swapping the order of the subtraction in \cref{eq:dist_weighted_4} does not change the value of the squared range used in \cref{eq:dist_weighted_5}. Combining \cref{eq:dist_weighted_8} and \cref{eq:dist_weighted_9}:
\begin{align}
    \prob_{S} (| \emprisk^*_{S}(h_i) - \risk^*_{\tilde{D}_S}(h_i)| \ge \epsilon ) \le 2e^{-2 N \epsilon^2 \prob(h_i)^2}.
\end{align}
Setting the right hand side to $\delta$ and solving for $\epsilon$ completes the derivation.
Namely for any $\delta > 0$, the following hold with probability $\le \delta$:
\begin{align}
    |\emprisk^*_{S}(h_i) - \risk^*_{\tilde{D}_S}(h_i)| \ge \frac{1}{\prob(h_i)} \sqrt{\frac{\log \frac{2}{\delta}}{2 N}} \label{eq:abs_bound} \\
    \risk^*_{\tilde{D}_S}(h_i) - \emprisk^*_{S}(h_i) \ge \frac{1}{\prob(h_i)} \sqrt{\frac{\log \frac{2}{\delta}}{2 N}}.
\end{align}
Hence the following hold with probability $> 1 - \delta$:
\begin{align}
    \risk^*_{\tilde{D}_S}(h_i) < \emprisk^*_{S}(h_i) + \frac{1}{\prob(h_i)} \sqrt{\frac{\log \frac{2}{\delta}}{2 N}} \\
    \risk^*_{\tilde{D}_S}(h_i) \leq \emprisk^*_{S}(h_i) + \frac{1}{\prob(h_i)} \sqrt{\frac{\log \frac{2}{\delta}}{2 N}}. \label{eq:emp_risk_smooth_bound}
\end{align}
\hfill$\square$.

%\Cref{thm:true_error} assumes dataset sizes for each activation region are constant, which holds asymptotically for large draws of the full dataset $S$, for fixed pre-trained f.

%\newpage
\section{Efficient computation of bound} \label{s:practical_derivations}

For normalization in \cref{eq:density}, we reduce computational complexity from exponential $\mathcal{O}(N2^{M})$ to linear $\mathcal{O}(M)$ (derivations below):
\begin{align}
    \sum_{l \in [1, C]} \sum_{\substack{j \in [1, C] \\ \text{s.t. } N_j > 0} } N_j  \spa k(h_l, h_j) = N \sum_{b = 0}^{M} d(b) \spa  {M \choose b} \eqdef u, \label{eq:practical_1}
\end{align}
and for normalization in \cref{eq:dist_weighted_fact4}, from exponential $\mathcal{O}(N^2 2^{M})$ to polynomial $\mathcal{O}(N^2 + M^3)$:
\begin{align}
    \forall_{\substack{l \in [1, C] \\ \text{s.t. } N_l > 0}}  &: \sum_{j \in [1, C]} \prob(h_j) \spa k(h_l, h_j) = \frac{1}{u}\sum_{\substack{i \in [1, C] \\ \text{s.t. } N_i > 0}} N_i \spa z(\operatorname{Hamming}(p_l, p_i)), \label{eq:practical_2_1} \\
    \text{where } \forall a \in [0, M] &: z(a) \eqdef \sum_{b=0}^{M} 
    d(b) \sum_{c=0}^{b}
    {a \choose c} 
    {M - a \choose b - c} \spa d(a + (b - c) - c ). \label{eq:practical_2_2}
\end{align}

\paragraph{Normalization term in \cref{eq:density}.} Computed once for the network.
\begin{align}
    \sum_{l \in [1, C]} \sum_{\substack{j \in [1, C] \\ \text{s.t. } N_j > 0} } N_j  \spa k(h_l, h_j) = \sum_{\substack{j \in [1, C] \\ \text{s.t. } N_j > 0}} N_j \sum_{l \in [1, C]} k(h_l, h_j) = N \sum_{b = 0}^{M} d(b) \spa  {M \choose b} \eqdef u, \label{eq:practical_deriv_1}
\end{align}
where the trick is that $\sum_{l \in [1, C]} k(h_l, h_j) = \sum_{l \in [1, C]} d(\operatorname{Hamming}(p_l, p_j))$ is the same for all $h_j$ due to completeness of the bit space $P$, and the specific value can be found by looping through all possible Hamming distances $b$.
%(number of bits flipped going from $p_j$ to $p_l$) and summing the k value with its count. 
%This enables computational complexity reduction 
This reduces the computational complexity 
from $\mathcal{O}(N2^{M})$ to $\mathcal{O}(M)$, because the number of populated activation regions, i.e. length of loop over $j$, is upper bounded by the number of samples $N$, and ${M \choose b}$ can be iteratively updated in $\mathcal{O}(1)$ in the loop over $b$:
\begin{align}
    {M \choose b} = {M \choose b-1} \frac{M - (b - 1)}{b}. \label{e:choose_iterative}
\end{align}

\paragraph{Normalization term in \cref{eq:dist_weighted_fact4}.} Computed $\forall l \in [1, C] \text{ s.t. } N_l > 0$:
%
%  h'' = h_l, h''' = h_j, h'''' = h_i
\vspace{-0.5em}
\begin{align}
    &\sum_{j \in [1, C]} \prob(h_j) \spa k(h_l, h_j)  \label{eq:practical_deriv_2_1} \\
    & = \frac{1}{u} \sum_{j \in [1, C]} \sum_{\substack{i \in [1, C] \\ \text{s.t. } N_i > 0}}  k(h_l, h_j) \spa  k(h_j, h_i) \spa N_i \label{eq:practical_deriv_2_2} \\
    &= \frac{1}{u}\sum_{\substack{i \in [1, C] \\ \text{s.t. } N_i > 0}} N_i \sum_{j \in [1, C]} k(h_l, h_j) \spa  k(h_j, h_i) \label{eq:practical_deriv_2_3}  \\
    &= \frac{1}{u}\sum_{\substack{i \in [1, C] \\ \text{s.t. } N_i > 0}} N_i 
    \overbrace{
    \sum_{b=0}^{M} \spa
    \overbrace{d(b)}^{\text{from } k(h_l, h_j)} \sum_{c=0}^{b}
    {\operatorname{Hamming}(p_l, p_i) \choose c} 
    {M - \operatorname{Hamming}(p_l, p_i) \choose b - c} \cdot %\hspace{1.5em}
    }
    ^{= z(\operatorname{Hamming}(p_l, p_i)) \text{ (\cref{eq:practical_2_2})}}
    \nonumber \\
    &\hspace{20.5em} \overbrace{d(\operatorname{Hamming}(p_l, p_i) + (b - c) - c )}^{\text{from } k(h_j, h_i)}. 
    \label{eq:practical_deriv_2_4} 
    % xor
    % choose from diffs, then choose from sames
\end{align}
%\vspace{-0.5em}
Note the loops over $i$ and $l$ only need to range over populated activation regions due to the inclusion of their sample counts ($N_i$ and $N_l$) as multiplicative factors in their loop contents, which means the iterations are bounded by $\mathcal{O}(N)$.
The problematic loop is over $j$, i.e. all subfunctions/activation regions whether populated by samples or not, which is $\mathcal{O}(2^{M})$ when $P$ is the full bit space $\{0, 1\}^{M}$. However, similar to \cref{eq:practical_deriv_1} above, it is also this fullness that we exploit.
%to group subfunctions $h_j$ by distance from $h_l$, i.e. $b$, which allows us to compute the sum over all $j$ cheaply by looping explicitly over all $b$ instead of $j$.

%\begin{wrapfigure}[12]{r}{0.4\textwidth}
\begin{wrapfigure}[]{r}{0.4\textwidth}
 \vspace{-1.5em}
%\begin{figure}[h]
    \centering
    \includegraphics[width=0.4\textwidth]{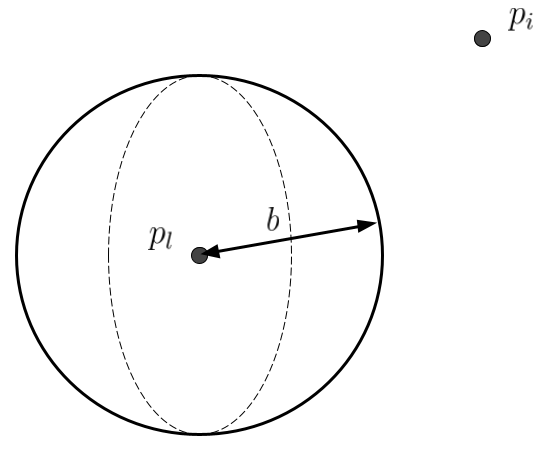}
    \caption{Sphere of $p_j$ that are $b$ bits away from $p_l$. For illustration only.
    %; activation pattern space $P$ is discrete binary and the metric producing $b$ is Hamming distance.
    }
     \vspace{-1em}
    \label{f:hamming}
%\end{figure}
\end{wrapfigure}
%\FloatBarrier

% hamming is number of different bits
% k and d are closeness (high with close), Hamming is distance (high with far)

Now we explain \cref{eq:practical_deriv_2_4} in detail. 
%First, recall that $k, d$ are high valued with closeness between patterns, and $\operatorname{Hamming}$ (or bit difference) is low valued with closeness between patterns.
The value $z(\operatorname{Hamming}(p_l, p_i))$ corresponds to the sum over $j$ in \cref{eq:practical_deriv_2_3}: loop over every bit pattern $p_j$ in $P$, multiply its closeness to $p_l$ with its closeness to $p_i$ (both fixed outside this loop), and sum.
Now group $p_j$ by bit distance to $p_l$, $b$. Consider one instance of $b$ (\cref{f:hamming}). Within this group for $b$, the closeness of every $p_j$ to $p_l$ is constant, namely equal to $d(b)$. But they have different distances to $p_i$. 
However we know the number of bits that are different between $p_l$ and $p_i$: this is $\operatorname{Hamming}(p_l, p_i)$.
Consider a specific instance of $p_j$.
Visualize $p_l$ turning into $p_j$ by flipping bit one at a time; each will either bring the pattern closer to or further away from $p_i$. There is a budget of $b$ different bits to flip to reach $p_j$. Say $c$ of those flips brought it closer (i.e. turned the bit value at that position into the same as $p_i$'s). Then we know $b - c$ brought it further.
So the number of different bits between $p_j$ and $p_i$ is exactly $\operatorname{Hamming}(p_l, p_i) + (b - c) - c$.
And the number of $p_j$ that fit this description is the number of ways of choosing $c$ from the different bits between $p_l$ and $p_i$ multiplied by the number of ways of choosing $b - c$ from the shared bits between $p_l$ and $p_i$: ${\operatorname{Hamming}(p_l, p_i) \choose c} 
{M - \operatorname{Hamming}(p_l, p_i) \choose b - c} $.

Conveniently,  ${q \choose g} = 0$ for $g > q$, so there is no need to add special cases to exclude the cases where the number we seek to choose is greater than the number available.

\Cref{eq:practical_deriv_2_4} enables computational complexity reduction of \cref{eq:practical_deriv_2_2}, including the outer loop over $l$, from exponential to polynomial time: $\mathcal{O}(N^2 2^{M})$ to $\mathcal{O}(N^2 + M^3)$ (excluding computation of $u$). The $M^3$ comes from constructing lookup table $z$, with the cubed power coming from looping over all possible values for $\operatorname{Hamming}(p_l, p_i)$, $b$ and $c$. This subsumes $M^2$ to compute $\forall q \in [0, M], \forall g \in [0, M]: {q \choose g}$ in the same iterative manner as \cref{e:choose_iterative}.

\newpage
\section{Pseudo-code}
\SetArgSty{textnormal}
\begin{algorithm}[h]
\footnotesize
\SetAlgoLined
    \textbf{Require:} pre-trained model $f_\theta$ and training/validation/test datasets. 
    
    \vspace{0.5em}
    
    Compute activation patterns for training and validation data from last ReLU layer;

    \For{\text{hyperparameter values } $\delta, \rho$}{
        Compute global normalization constant (\cref{eq:practical_1});
        
        For training data activation patterns, compute normalization constants (\cref{eq:practical_2_1});
        
        For validation data activation patterns, compute $\operatorname{log}$ bound using $\delta, \rho$ and normalization constants (\cref{eq:true_error_bound});
        
        Compute validation metric (AUCEA) with $\operatorname{log}$ bound of sample's activation pattern as unreliability;
        
        \If{highest validation metric}{
           Store $\delta, \rho$ as best with normalization constants;
        }
    }
    
    \vspace{0.5em}

    Compute activation patterns for test data from last ReLU layer;
    
    For test data activation patterns, compute $\operatorname{log}$ bound using chosen $\delta, \rho$ and normalization constants (\cref{eq:true_error_bound});
        
    Compute test metrics with $\operatorname{log}$ bound of sample's activation pattern as unreliability.
    
 \caption{Subfunction error bounds for predicting sample prediction unreliability}
 \label{a:algo}
\end{algorithm}

\FloatBarrier

\section{Additional results tables}

\begin{table}[h]
    \footnotesize
    \centering
    \caption{Test accuracy of original models.}
    \begin{tabular}{c c c c} 
\toprule
Dataset & Model & Accuracy \\
\midrule
CIFAR100 & VGG16 & 0.704 $\pm$ 1E-03 \\
CIFAR100 & ResNet50 & 0.729 $\pm$ 8E-03 \\
CIFAR10 & VGG16 & 0.920 $\pm$ 2E-03 \\
CIFAR10 & ResNet50 & 0.908 $\pm$ 5E-03 \\
\bottomrule
    \end{tabular}
    \label{t:orig_acc}
\end{table}

\begin{table}[h]
    \centering
    \footnotesize
    \caption{Model trained on CIFAR100. Out-of-distribution detection (AUROC) vs CIFAR10/SVHN. }
    \label{t:ood_cifar100}
    \begin{tabular}{l c c c c}
\toprule
 & \multicolumn{2}{c}{$\rightarrow$ CIFAR10}  & \multicolumn{2}{c}{$\rightarrow$ SVHN} \\
\midrule
 & VGG16  & ResNet50  & VGG16  & ResNet50 \\
\midrule
Residual flows density & 0.495 $\pm$ 2E-04  & 0.495 $\pm$ 2E-04  & 0.090 $\pm$ 2E-04  & 0.090 $\pm$ 2E-04 \\
GP & 0.708 $\pm$ 6E-03  & 0.404 $\pm$ 3E-02  &  \textbf{0.830 $\pm$ 3E-02 }  & 0.396 $\pm$ 8E-02 \\
Class distance & 0.627 $\pm$ 4E-02  & 0.513 $\pm$ 4E-02  &  \textbf{0.833 $\pm$ 3E-02 }  & 0.579 $\pm$ 1E-01 \\
Margin & 0.716 $\pm$ 2E-03  & 0.736 $\pm$ 3E-03  & 0.771 $\pm$ 5E-03  & 0.790 $\pm$ 3E-02 \\
Entropy &  \textbf{0.725 $\pm$ 3E-03 }  & 0.745 $\pm$ 3E-03  & 0.786 $\pm$ 6E-03  &  \textbf{0.814 $\pm$ 4E-02 } \\
Max response & 0.719 $\pm$ 3E-03  & 0.740 $\pm$ 3E-03  & 0.776 $\pm$ 6E-03  & 0.799 $\pm$ 3E-02 \\
MC dropout & 0.693 $\pm$ 3E-03  & 0.725 $\pm$ 4E-03  & 0.771 $\pm$ 1E-02  & 0.795 $\pm$ 3E-02 \\
Cluster distance & 0.639 $\pm$ 1E-02  &  \textbf{0.754 $\pm$ 4E-03 }  & 0.561 $\pm$ 2E-02  &  \textbf{0.810 $\pm$ 5E-02 } \\
Subfunctions (ours) &  \textbf{0.738 $\pm$ 2E-03 }  &  \textbf{0.750 $\pm$ 7E-03 }  & 0.797 $\pm$ 8E-03  & 0.807 $\pm$ 3E-02 \\
\bottomrule
    \end{tabular}
\end{table}

\begin{table}[h]
    \centering
    \footnotesize
    \caption{Model trained on CIFAR100. Predicting misclassification on in-distribution test (AUCEA and AUROC).}
    \label{t:id_cifar100}
    \begin{tabular}{l c c c c}
\toprule
 & \multicolumn{2}{c}{VGG16}  & \multicolumn{2}{c}{ResNet50} \\
\midrule
 & AUCEA  & AUROC  & AUCEA  & AUROC \\
\midrule
Residual flows density & 0.293 $\pm$ 1E-02  & 0.622 $\pm$ 1E-02  & 0.293 $\pm$ 2E-02  & 0.630 $\pm$ 1E-02 \\
GP & 0.882 $\pm$ 1E-03  & 0.803 $\pm$ 2E-03  & 0.670 $\pm$ 2E-02  & 0.384 $\pm$ 2E-02 \\
Class distance & 0.838 $\pm$ 3E-02  & 0.739 $\pm$ 6E-02  & 0.627 $\pm$ 7E-02  & 0.476 $\pm$ 3E-02 \\
Margin & 0.899 $\pm$ 2E-03  & 0.852 $\pm$ 4E-03  & 0.870 $\pm$ 6E-03  & 0.855 $\pm$ 4E-03 \\
Entropy & 0.895 $\pm$ 2E-03  &  \textbf{0.856 $\pm$ 5E-03 }  &  \textbf{0.916 $\pm$ 4E-03 }  &  \textbf{0.859 $\pm$ 4E-03 } \\
Max response &  \textbf{0.899 $\pm$ 2E-03 }  & 0.853 $\pm$ 4E-03  & 0.864 $\pm$ 7E-03  &  \textbf{0.857 $\pm$ 4E-03 } \\
MC dropout & 0.894 $\pm$ 1E-03  & 0.828 $\pm$ 8E-03  & 0.898 $\pm$ 5E-03  & 0.841 $\pm$ 2E-03 \\
Cluster distance & 0.833 $\pm$ 9E-03  & 0.722 $\pm$ 2E-02  & 0.900 $\pm$ 5E-03  & 0.824 $\pm$ 7E-03 \\
Subfunctions (ours) &  \textbf{0.904 $\pm$ 1E-03 }  &  \textbf{0.864 $\pm$ 3E-03 }  &  \textbf{0.902 $\pm$ 4E-03 }  & 0.827 $\pm$ 4E-03 \\
\bottomrule
\end{tabular}
\end{table}

\FloatBarrier
\section{Metrics}\label{app:metrics}
Evaluation metrics were area under the graphs of coverage vs. effective accuracy (AUCEA) and false positive rate vs. true positive rate (AUROC), with the latter as standard for OOD experiments. 

\begin{align}
    \text{coverage} = \frac{TP + FP}{TP + FP + TN + FN} \qquad  \text{ef}&\text{fective accuracy} = \text{precision} = \frac{TP}{TP + FP} \\
    \text{false positive rate} = \frac{FP}{FP + TN} \qquad &\text{true positive rate} = \frac{TP}{TP + FN}.
\end{align}

\section{Additional experimental details} \label{app:exp}

Experiments were averaged over 5 random seeds. The classification models were trained with SGD optimization with learning rate 0.1, momentum 0.9, weight decay 5e-4 and standard schedules: 100 epochs with learning rate $^*0.1$ every 30 epochs (CIFAR10) and 200 epochs with learning rate $^*0.2$ every 60 epochs (CIFAR100). 
For each unreliability metric, each threshold yielded one set of accept/reject decisions for the test set which yielded one point in each evaluation graph, and area under graph was computed using the trapezoidal rule. 
For the in-distribution setting, AUCEA corresponds to effective model accuracy (i.e. of accepted samples) averaged over different thresholds, and thus can be compared against original model accuracy.

\subsection{Subfunction error bound hyperparameters}

The searched values for likelihood parameter $\delta$ were [0.001, 0.01, 0.1, 0.2, 0.3, 0.4, 0.5, 0.6, 0.7, 0.8, 0.9] and for weighting function standard deviation parameter $\rho$ were [32, 48, 64, 98, 128, 192, 256, 384, 512] (VGG16) and [4, 6, 8, 12, 16, 24, 32, 48, 64] (ResNet50). Validation set AUCEA was used for selection. The selected values are in \cref{tab:hyperparams}.

\begin{table}[h]
    \centering
        \caption{Hyperparameters used for subfunction error. Brackets denote number of seeds.}
    \begin{tabular}{c c c c}
        \toprule
         Dataset & Model & $\rho$  & $\delta$ \\
         \midrule
CIFAR100 & VGG16  &  128.0 (\#: 5) &  0.001 (\#: 5)\\
CIFAR100 & ResNet50  &  16.0 (\#: 5) &  0.001 (\#: 4), 0.1 (\#: 1)\\
CIFAR10 & VGG16  &  48.0 (\#: 3), 98.0 (\#: 1), 64.0 (\#: 1) &  0.001 (\#: 4), 0.9 (\#: 1)\\
CIFAR10 & ResNet50  &  16.0 (\#: 3), 24.0 (\#: 2) &  0.001 (\#: 4), 0.3 (\#: 1)\\
         \bottomrule
    \end{tabular}
    \label{tab:hyperparams}
\end{table}

\subsection{Dataset statistics} \label{s:dataset_stats}
All results infer unreliability of the test sets. The datasets are publicly available.
\begin{table}[h]
\centering
\begin{tabular}{c c c c}
    \toprule
    Dataset & Train & Validation & Test \\
    \midrule
     CIFAR10 \citep{cifar} & 42500 & 7500 & 10000 \\
     CIFAR100 \citep{cifar} & 42500 & 7500 & 10000 \\
     SVHN \citep{netzer2011reading} & 62269 & 10988 & 26032 \\
     \bottomrule
\end{tabular}
\end{table}

\subsection{Computational resources}\label{s:compute}

Experiments were run given a shared cluster of machines with approximately 140 GPUs. Jobs required less than 24GB GPU memory.
For subfunction error, normalization constants were computed first in a pre-computation phase.
This took up most of the runtime and was parallelized by splitting jobs by architecture, dataset, seed and $\rho$; each job took approximately 20 minutes. 
Subsequent inference on the test sets (i.e. computing unreliability of unseen samples) took approximately 2 - 10 minutes per combination of architecture, dataset and seed.

\newpage
\section{Additional Boxplots for other methods}\label{app:boxplots}

\begin{wraptable}{r}{61mm}
\vspace{-1em}
%\begin{table}
\centering
\footnotesize
\caption{Pearson correlation coefficients between unreliability rank and semantic novelty w.r.t. CIFAR10, on CIFAR100 data and CIFAR10 model. Higher is better.}\label{t:boxplot_coeffs}
\begin{tabular}{l c}
\toprule
Method & Correlation \\
\midrule
Residual flows density  & -0.259 \\
GP &  0.390 \\
Class distance &   0.171 \\
Margin & 0.231 \\
Entropy & 0.252 \\
Max response & 0.242 \\
MC dropout & 0.433 \\
Cluster distance & 0.347 \\
\textbf{Subfunctions (ours)} & \textbf{0.511} \\
\bottomrule
\end{tabular}
%\end{table}
\end{wraptable}

Settings apart from the method are the same as \cref{fig:boxplot}. We measured the Spearman correlation coefficient between semantic novelty (green=0, purple=1) and unreliability (median unreliability rank for each class) and found the correlation was lower for all baselines compared to subfunctions, which had a correlation coefficient of 0.511. The closest baseline method was MC dropout (\cref{t:boxplot_coeffs}).

\vspace{9em}

\begin{figure}[h]
    \centering
    \includegraphics[width=\textwidth]{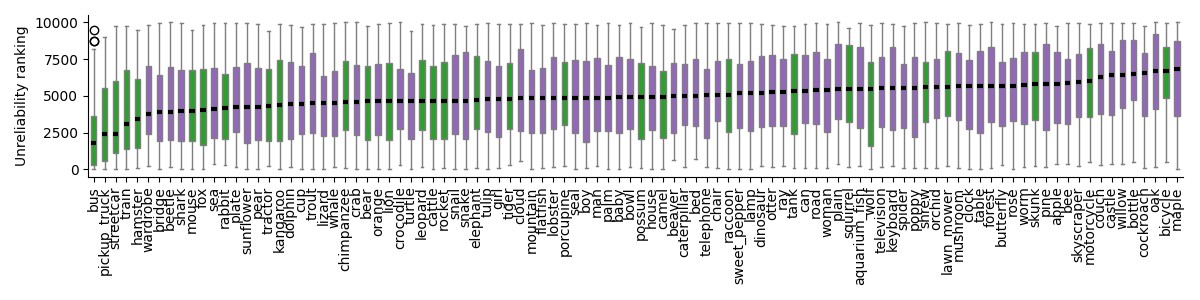}
    \caption{Entropy.}
\end{figure}

\begin{figure}[h]
    \centering
    \includegraphics[width=\textwidth]{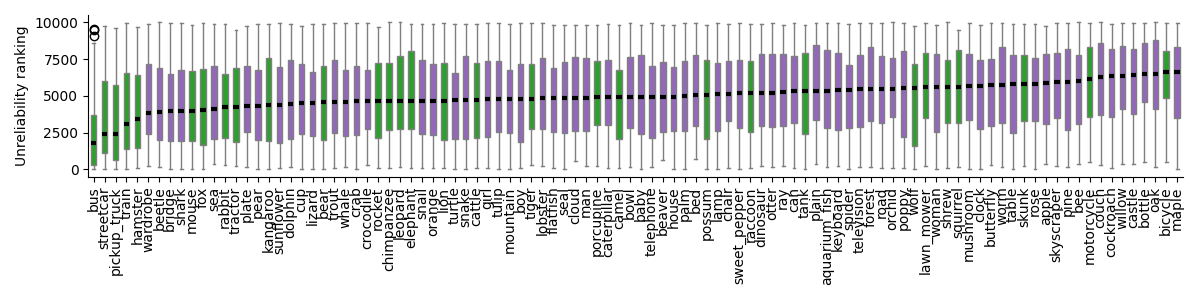}
    \caption{Max response.}
\end{figure}

\begin{figure}[h]
    \centering
    \setlength{\abovecaptionskip}{-0.5em}
    \includegraphics[width=\textwidth]{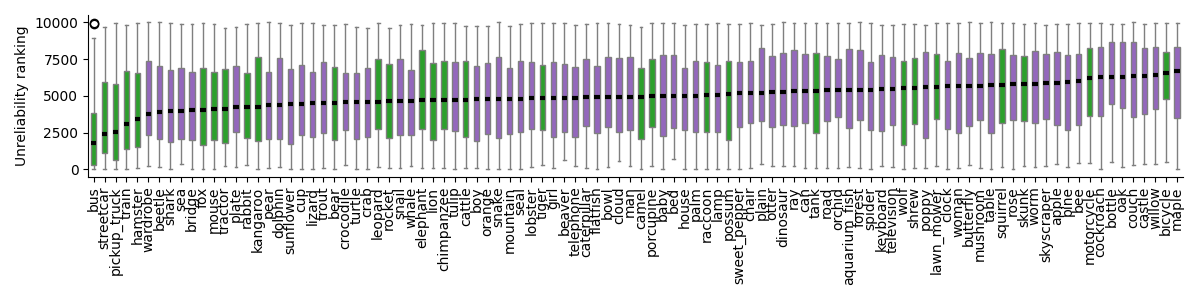}
    \caption{Margin.}
\end{figure}

\begin{figure}
    \centering
    \includegraphics[width=\textwidth]{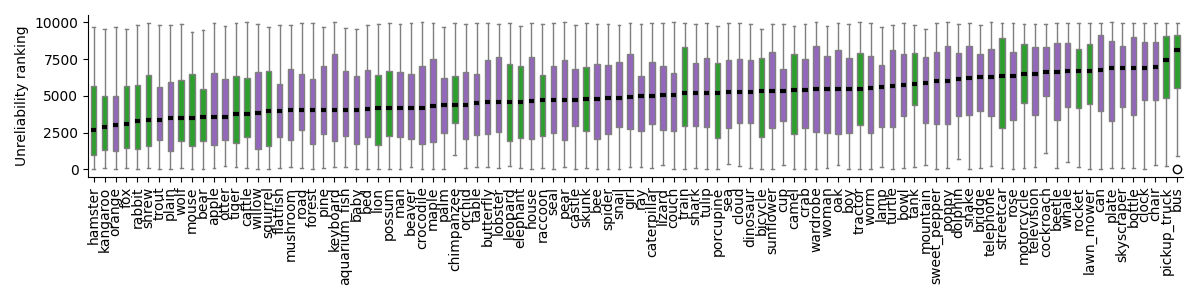}
    \caption{Class distance.}
\end{figure}

\begin{figure}
    \centering
    \includegraphics[width=\textwidth]{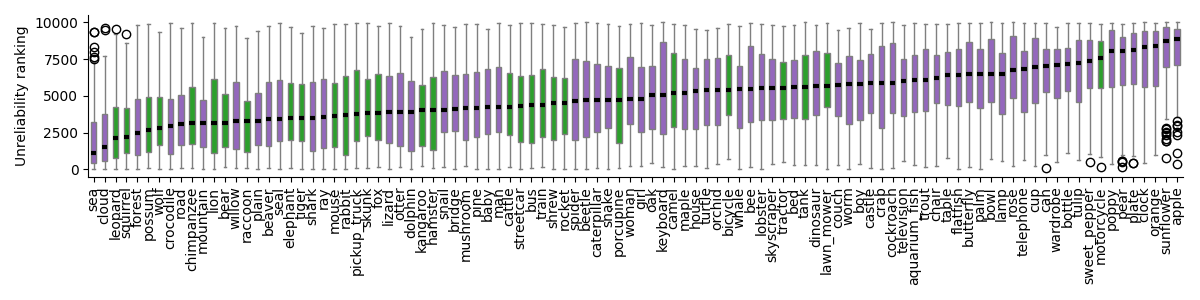}
    \caption{GP.}
\end{figure}

\begin{figure}
    \centering
    \includegraphics[width=\textwidth]{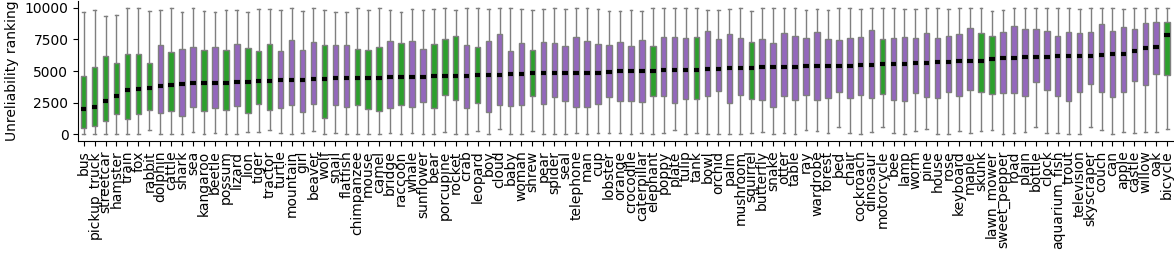}
    \caption{MC dropout.}
\end{figure}

\begin{figure}
    \centering
    \includegraphics[width=\textwidth]{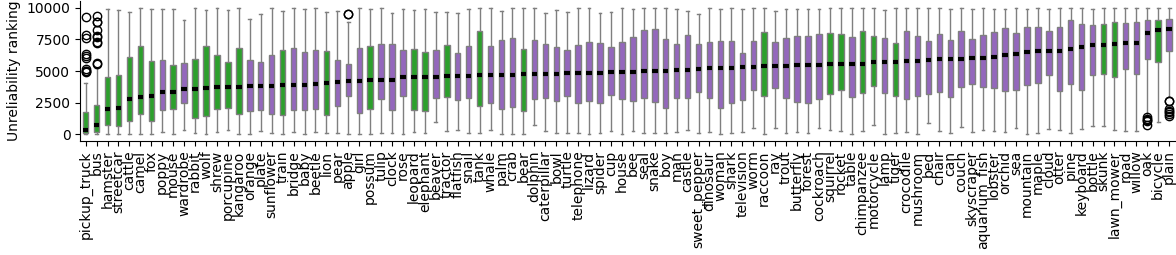}
    \caption{Cluster distance.}
\end{figure}

\begin{figure}
    \centering
    \includegraphics[width=\textwidth]{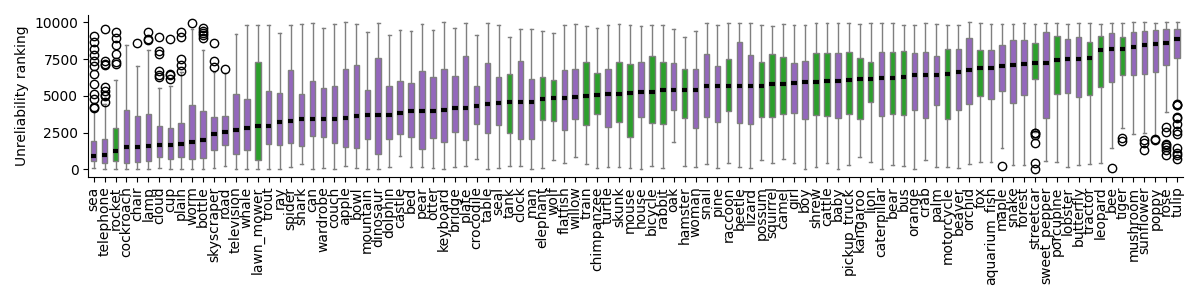}
    \caption{Residual flows density.}
\end{figure}

\end{document}